\documentclass{article}

\usepackage{PRIMEarxiv}

\usepackage[utf8]{inputenc} 
\usepackage[T1]{fontenc}    
\usepackage{hyperref}       
\usepackage{url}            
\usepackage{booktabs}       
\usepackage{amsfonts}       
\usepackage{nicefrac}       
\usepackage{microtype}      
\usepackage{lipsum}
\usepackage{fancyhdr}       
\usepackage{graphicx}       
\graphicspath{{media/}}     

\title{MapAI: Precision in Building Segmentation}

\author{
  Sander Riisøen Jyhne \\
  The Norwegian Mapping Authority \\
  University of Agder \\
  Grimstad, Norway \\
\texttt{sander.jyhne@kartverket.no} \\
\And
Morten Goodwin \\
  Centre for Artificial Intelligence Research \\
  University of Agder \\
  Grimstad, Norway \\
\texttt{morten.goodwin@uia.no} \\
\And
Per Arne Andersen \\
  Centre for Artificial Intelligence Research \\
  University of Agder \\
  Grimstad, Norway \\
\texttt{per.andersen@uia.no} \\
\And
Ivar Oveland \\
  The Norwegian Mapping Authority \\
  Kristiansand, Norway \\
\texttt{morten.goodwin@uia.no} \\
\And
Alexander Salveson Nossum \\
  Norkart \\
  Kristiansand, Norway \\
\And
Karianne Ormseth \\
  AI:Hub \\
  Grimstad, Norway \\
\And
Mathilde Ørstavik \\
  Norkart \\
  Oslo, Norway \\
\And
Andrew C. Flatman \\
  SDFI \\
  Denmark \\}

\begin{document}
\maketitle

\begin{abstract}
\textit{MapAI: Precision in Building Segmentation} is a competition arranged with the Norwegian Artificial Intelligence Research Consortium (NORA) \footnote{\url{https://nora.ai}} in collaboration with Centre for Artificial Intelligence Research at the University of Agder (CAIR)\footnote{\url{https://cair.uia.no}}, the Norwegian Mapping Authority\footnote{\url{https://kartverket.no}}, AI:Hub\footnote{\url{https://aihub.no/}}, Norkart\footnote{\url{https://www.norkart.no}}, and the Danish Agency for Data Supply and Infrastructure\footnote{\url{https://eng.sdfi.dk/}}. The competition will be held in the fall of 2022. It will be concluded at the \textit{Northern Lights Deep Learning conference} focusing on the segmentation of buildings using aerial images and laser data. We propose two different tasks to segment buildings, where the first task can only utilize aerial images, while the second must use laser data (LiDAR) with or without aerial images. Furthermore, we use IoU and Boundary IoU \cite{biou} to properly evaluate the precision of the models, with the latter being an IoU measure that evaluates the results' boundaries. We provide the participants with a training dataset and keep a test dataset for evaluation.


\end{abstract}


\section{Introduction}
Buildings are an essential component of information regarding population, policy-making, and city management \cite{cityplanning}. Using computer vision technologies such as classification, object detection, and segmentation has proved helpful in several scenarios, such as urban planning and disaster recovery \cite{cityplanning, disaster_recovery}. Segmentation is the most precise method and can give detailed insights into the data as it highlights the area of interest.

Acquiring accurate segmentation masks of buildings is challenging since the training data derives from real-world photographs. As a result, the data often have varying quality, large class imbalance, and contains noise in different forms. The segmentation masks are affected by optical issues such as shadows, reflections, and perspectives. Additionally, trees, powerlines, or even other buildings may obstruct the visibility \cite{visibility}. Furthermore, small buildings have proved to be more difficult to segment than larger ones as they are harder to detect, more prone to being obstructed, and often confused with other classes \cite{smallobjects}. Lastly, different buildings are found in several diverse areas, ranging from rural to urban locations. The diversity poses a vital requirement for the model to generalize to the various combinations. These hardships motivate the competition and our evaluation method, detailed in \nameref{sec:evaluation}.

This competition follows mostly the same strategy as previous NORA competitions \cite{hicks2021medai,nordmo2022fish}. 

\section{Dataset Details}
For the competition, we provide the participants with a dataset containing aerial images, laser data, and ground truth masks for the buildings. We split the dataset into a training dataset and a test dataset. The training dataset is released at the start of the competition, while the test dataset will be kept hidden until the competition is over. When the competition is complete, we will release the full dataset.

The training dataset consists of several different locations in Denmark. Area variability ensures a diverse dataset with several different environments and building types. The test dataset consists of seven locations in Norway, comprising urban and rural cities.

The data is derived from real-world data. As a result, there are cases where the buildings in the aerial image do not correspond to a ground truth mask. In addition, the ground truths in the test dataset are generated using a DTM, which skews the top of the buildings in images compared to the ground truths. The training dataset is generated using a DSM, which does not skew the top of the buildings. In Figure \ref{fig:dataexample}, we can see examples of the different datatypes present in the training dataset.

\begin{figure}[h]
\centering
    \includegraphics[width=0.4\textwidth]{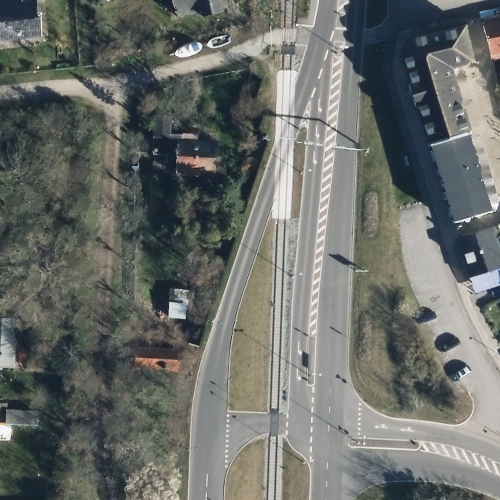}
    \caption{Aerial image sample from the training dataset}
    \label{fig:first}
\end{figure}
\begin{figure}[h]
\centering
    \includegraphics[width=0.4\linewidth]{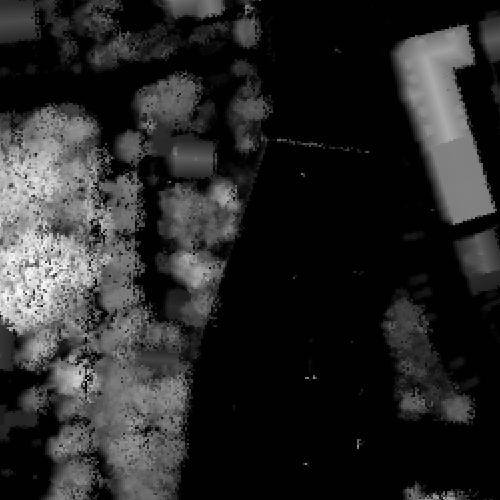}
    \caption{Lidar sample from the training dataset}
    \label{fig:second}
\end{figure}
\begin{figure}[h]
\centering
    \includegraphics[width=0.4\linewidth]{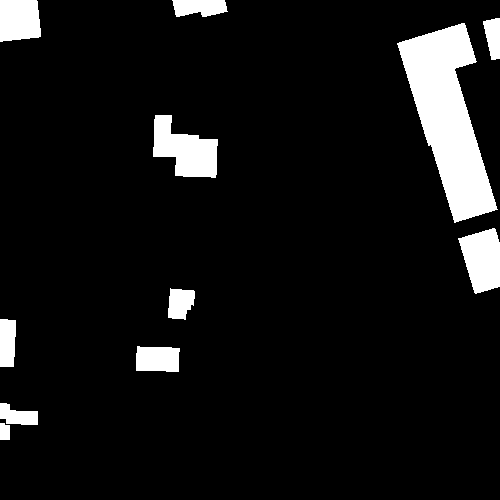}
    \caption{Mask sample from the training dataset}
    \label{fig:third}
\end{figure}
        
\section{Task Descriptions}
We present two subtasks; (1) an aerial image segmentation task and (2) laser data segmentation task. The participants are encouraged to submit for both tasks; however, it is not mandatory. Note that both tasks contribute equal weight to the final score, i.e., 50\% each. You can only reach the maximum score by a separate submission for both tasks

\subsection{Task 1: Aerial Image Segmentation Task}
The aerial image segmentation task aims to solve the segmentation of buildings only using aerial images. Segmentation using only aerial images is helpful for several scenarios, including disaster recovery in remote sensing images where laser data is unavailable. We ask the participants to develop machine learning models for generating accurate segmentation masks of buildings solely using aerial images.

\subsection{Task 2: Laser Data Segmentation Task}
The laser data segmentation task aims to solve the segmentation of buildings using laser data. Segmentation using laser data is helpful for urban planning or change detection scenarios, where precision is essential. We ask the participants to develop machine learning models for generating accurate segmentation masks of buildings using laser data with or without aerial images.

\subsection{Submission}
We have developed a new template for competition participation and submission. The contestants will have to fork a GitHub repository, create a folder for their team and develop their methods in the folder. When the participants are ready for submission, they push the changes to their fork and create a pull request to the original repository. Automatic Github actions will test their code and models and store their results. The GitHub repository and instructions are found at \url{https://github.com/Sjyhne/mapai-competition}.

\section{Evaluation Methodology}
\label{sec:evaluation}
We evaluate both tasks with the same metrics described in the following paragraphs. Overall there will be a first place and a second place winner of MapAI, determined by the sum of the score $S$ from both tasks.
\newline
\newline
We use a combination of Intersection-over-Union (IoU) and Boundary Intersection-over-Union (BIoU) to evaluate the submitted segmentation mask. The formula for the evaluation score $S$ is presented in Eq. \ref{math:score}.


\begin{equation}
\label{math:score}
    S = \frac{BIoU + IoU}{2}
\end{equation}

The Intersection-over-Union (IoU), also known as the Jaccard Index, measures the similarity between two samples, \textit{G} (ground truth) and \textit{P} (prediction), by dividing the intersecting area by the total area, seen in Eq. \ref{math:iou}.

\begin{equation}
\label{math:iou}
    IoU = \frac{Intersection}{Union} = \frac{|G \cap P|}{|G| + |P| - |G \cap P|}
\end{equation}

\begin{figure}[h]
    \includegraphics[width=0.6\linewidth]{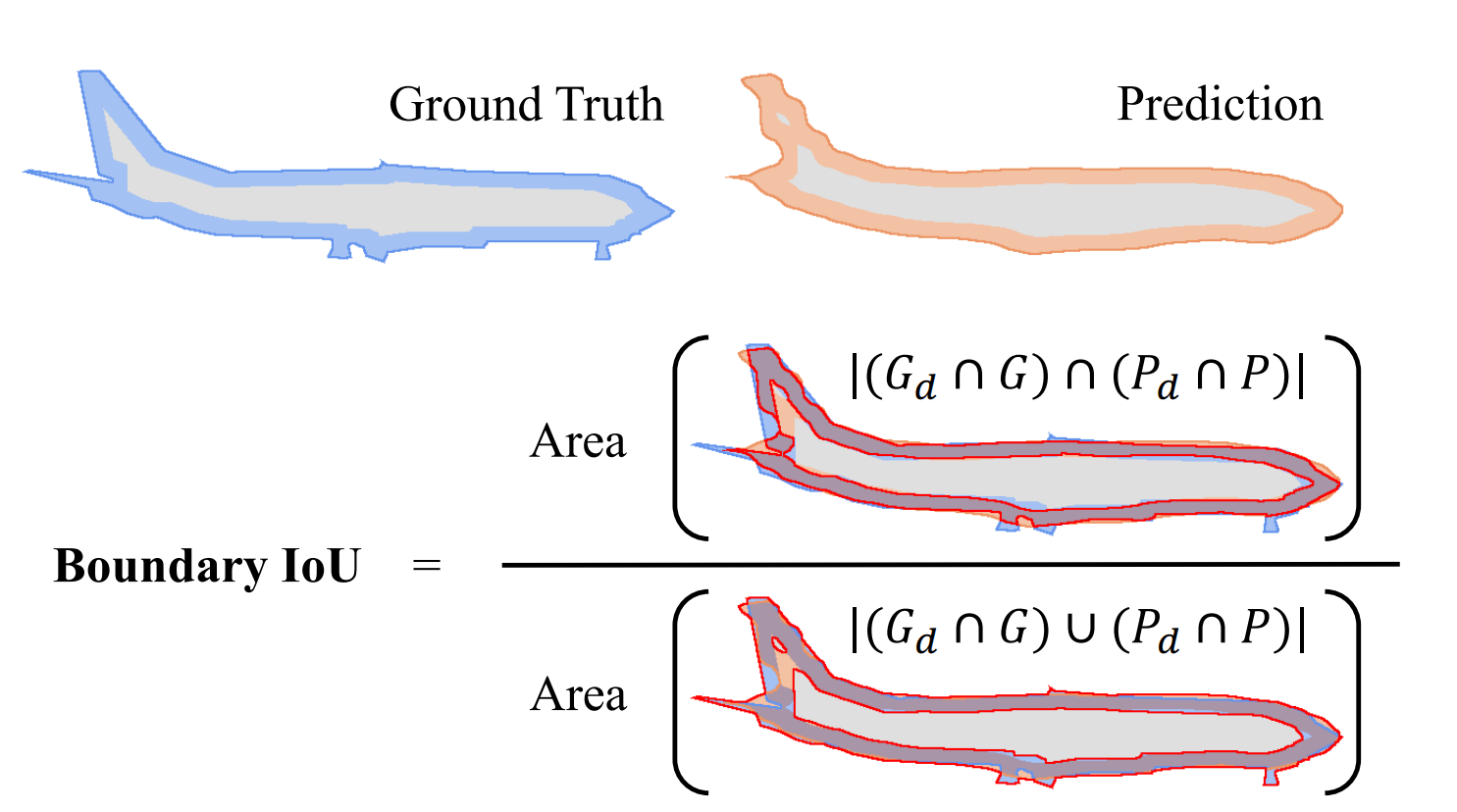}
    \centering
    \caption{The Boundary Intersection-over-Union (BIoU) used to measure the accuracy of the segmentation boundary \cite{biou}}
    \label{fig:biou}
\end{figure}

The Boundary Intersection-over-Union (BIoU) calculates the IoU of the boundary of the prediction and ground truth. A variable \textit{d} determines the width of the boundary used in the calculation. Fig \ref{fig:biou} visualizes how we calculate the Boundary IoU, where $G$ and $G_d$ denote the ground truth and the edge of the ground truth with thickness \textit{d}. Similarly to \textit{G}, $P$ and $P_d$ denote the predicted mask and the edge of the predicted mask with thickness \textit{d}.

The final score for the competition is the average of the score $S$ from both tasks.

\section{Summary}
This paper presents the first version of \textit{MapAI: Precision in Building Segmentation} task. The competition aims to advance the remote sensing image segmentation field by presenting two subtasks with an evaluation focused on small buildings and segmentation boundaries. We hope the competition motivates students to participate and contribute to advancing the field.

\bibliographystyle{unsrt}  
\bibliography{references}

\end{document}